\documentclass[conference]{IEEEtran}
\IEEEoverridecommandlockouts
\usepackage{cite}
\usepackage{amsmath,amssymb,amsfonts}
\usepackage{algorithmic}
\usepackage{graphicx}
\usepackage{subcaption}
\usepackage{textcomp}
\usepackage{xcolor}
\def\BibTeX{{\rm B\kern-.05em{\sc i\kern-.025em b}\kern-.08em
    T\kern-.1667em\lower.7ex\hbox{E}\kern-.125emX}}
\begin{document}

\title{Dual-Triplet Metric Learning for Unsupervised Domain Adaptation in Video-Based Face Recognition \\
}

\author{\IEEEauthorblockN{George Ekladious, Hugo Lemoine, Eric Granger }
\IEEEauthorblockA{\textit{Laboratoire d'imagerie, de vision et d'intelligence artificielle (LIVIA)} \\
\textit{Dept. of Systems Engineering, \'Ecole de technologie sup\'erieure}\\
Montreal, Canada \\
\{george.ekladious, eric.granger\}@etsmtl.ca \\ 
hugo.lemoine-st-andre.1@ens.etsmtl.ca}
\and
\IEEEauthorblockN{Kaveh Kamali, Salim Moudache}
\IEEEauthorblockA{\textit{Nuvoola Inc.} \\
Montreal, Canada \\
\{kaveh.kamali, salim.moudache\}@Nuvoola.com}
}

\maketitle

\begin{abstract}
The scalability and complexity of deep learning models remains a key issue in many of visual recognition applications like, e.g., video surveillance, where  fine tuning  with labeled image data from each new camera is required to reduce the domain shift between videos captured from the source domain, e.g., a laboratory setting, and the target domain, i.e, an operational environment.   
%
In many video surveillance applications, like face recognition (FR) and person re-identification, a pair-wise matcher is used to assign a query image captured using a video camera to the corresponding reference images in a gallery. The different configurations and operational conditions of video cameras can introduce significant shifts in the pair-wise distance distributions, resulting in degraded recognition performance for new cameras.  
In this paper, a new deep domain adaptation (DA) method is proposed to adapt the CNN embedding of a Siamese network using unlabeled tracklets captured with a new video cameras. To this end, a dual-triplet loss is introduced for metric learning, where two triplets are constructed using video data from a source camera, and a new target camera.  In order to constitute the dual triplets, a mutual-supervised learning approach is introduced where the source camera acts as a teacher, providing the target camera with an initial embedding. Then, the student relies on the teacher to iteratively label the positive and negative pairs collected during, e.g., initial camera calibration. Both source and target embeddings continue to simultaneously learn such that their pair-wise distance distributions become aligned. 
For validation, the proposed metric learning technique is used to train deep Siamese networks under different training scenarios, and is compared to state-of-the-art techniques for still-to-video FR on the COX-S2V and a private video-based FR dataset.  Results indicate that the proposed method can provide a level of accuracy that is comparable to the upper bound performance, in training scenario where labeled target data is employed to fine-tune the Siamese network.
\end{abstract}


\section{Introduction}

Learning  discriminant representations from face images and efficiently calibrating these embedding to new capturing devices and operational environments  is required for a wide range of biometric and surveillance applications such as, face recognition (FR) in surveillance systems, character identification and clustering for video captioning,  web search, etc. Representation learning methods from still face images are extensively studied where availability of extremely large datasets of still images facilitates deep learning methods and achieve human-level performance like in the current FR systems \cite{Schroff15}.

Learning video-based face representations, on the other hand, are harder to learn for two main reasons. First, face images are extracted from videos with unconstrained capturing conditions and that can introduce significant variability in facial appearances according to pose, illumination, scale, expression, etc. Second, the  publicly available video-based face datasets are of much less size, compared to the still face datasets, so it can be insufficient to train reliable deep representations. For instance,  large-scale labeled video database publicly available to date, such as YouTube face dataset \cite{Wolf11},  contains 3.4K videos in total from 1.5K different subjects, as apposed to the still VGG face2 dataset with 3.3M faces from 9K subjects \cite{Cao18}. 

One way to tackle the above challenges is to reduce the variability in the video-based face images, so they  become similar to the still images, and hence,  the powerful still-based face representations   become usable. For instance, autoencoder networks are used  to learn discriminant face embedding, and to reconstruct high-quality images (frontal, well-illuminated, less blurred faces with neutral expression) from video face images  captured  under various conditions \cite{Parchami17c}.  Such methods may require enough data from the target domain and can be impractical to calibrate new video sources using low-shot calibration data.

Another way to tackle the video-based face representation learning challenge  is to  generate enough video-based data from the large still data (by adding effects to the still images similar to these produced by the video capturing conditions) and then use the generated data to design new representations for the video-based face applications. For instance, artificially blur training data are generated to  account for a shortfall in real-world video training data. Using training data composed of both still images and artificially blurred data, a deep Convolutional Neural Network (CNN) is encouraged to learn blur insensitive features automatically \cite{Ding18}.  Domain-Specific Face Synthesis methods are proposed where the domain specific variations, e.g., pose, illumination, etc.,   are  projected onto the reference still of each individual, so that they  resemble individuals of interest under the capture conditions relevant to the operational domain \cite{Mokhayeri2019} \cite{hong2017}.  Such methods can be too complex to calibrate new video sources efficiently and may not cover the complete range of capturing conditions that can occur during operation.  

Recently, the video-based face representation learning is approached from the  Domain Adaptation (DA) perspective, where images from specific environment and capturing conditions  (mostly high quality still images) are considered as the source data and face images captured from a different environment and capturing conditions (mostly video-based images) are considered as the target data. Some methods used large labeled data from target domain  to fine tune an initial source model \cite{Wen2018}. More practical methods employed Unsupervised Domain Adaptation (UDA) to transfer the discriminant source model to the target data using unlabeled target data \cite{Wen2018}\cite{Luo2018}\cite{ganin2014}. These methods fit more the classical "single stream" networks and they are not designed specifically for the multiple-stream networks, e.g., the Siamese structure, that is employed for designing pair-wise face matchers, e.g., in Still to Video (S2V) face recognition,  based on deep metric learning. 

Some works on UDA  for deep distance metric learning were recently proposed \cite{Laradji2018}\cite{sohn2018}.  These method are  either not applicable to face representation learning since they are designed for closed- and small-set  problems, such as handwritten digit recognition \cite{Laradji2018}, or they  apply mixture of techniques to support more complex applications, but this comes with the expense of complexity that might hinder practical applications such as, automatic calibration of surveillance cameras \cite{Sohn2017}\cite{hong2017}. More recently, self-supervised Learning  approaches are proposed to automatically label  target data by leveraging  temporal and contextual information exist in videos, e.g., tracklets \cite{Sharma19}\cite{Wu13}\cite{Cinbis15}.  These methods  require abundant of unlabeled target data and availability of co-occurring tracklets (tracklets from different subjects in the same scene that are produced by accurate face detectors and trackers). 

This paper addresses the aforementioned limitations of adapting a video-based face representation for  a new video source or environment, so that the adaptation is possible given unlabeled target data or where the detection and tracking information are not accurate or informative to produce labeled tracklets.  The contributions of this work are as follows:

\begin{itemize}
 \item we introduce a new domain adaptation framework called Dual-Triplet Metric Learning (DTML) that applies a novel dual-triplet loss function and a new mutual supervised learning approach. The proposed framework allows for adapting deep pair-wise matchers to different domains by aligning their distance distributions. 
 \item we propose a mutual-supervised learning approach  where the source (teacher) iteratively  labels  the unlabeled target (student) data. 
 \item we applied the proposed dual-triplet loss and mutual-supervised learning approach to the still-to-video face recognition application and provided accuracy that is comparable to the state-of-the-art  methods for video face representation  learning, but with unsupervised domain adaptation capability.
 \end{itemize}
 
It is important to mention that the proposed framework can be applied to different modalities, while in this paper we assess the method using the video-based face recognition as a specific use case.

\section{Related Work}
\begin{figure*}
\begin{center}
\framebox(490,230){ \includegraphics[scale=0.34]{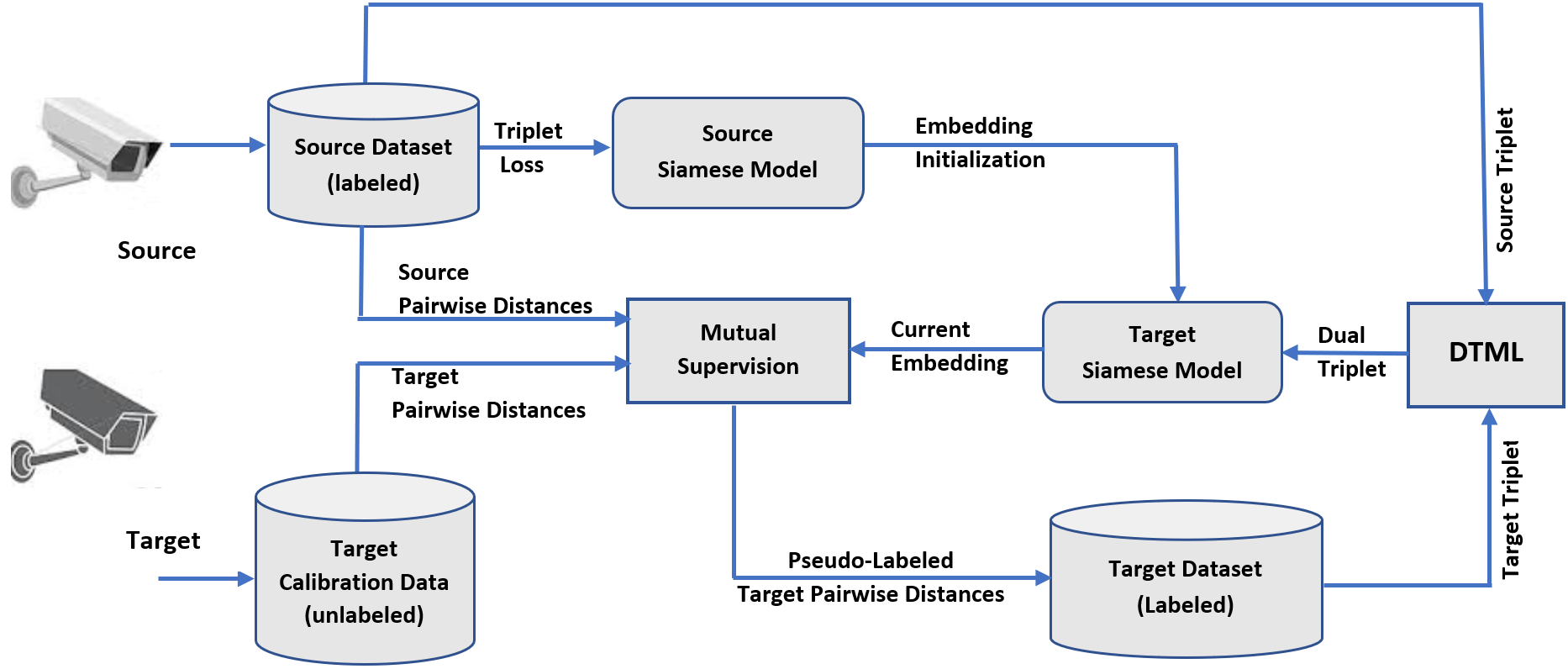}}

\end{center}
   \caption{Dual-Triplet Metric Learning for Domain Adaptation}
\label{fig:DTML}
\end{figure*}

This work has similarities to some  Unsupervised Domain Adaptation (UDA) methods for deep metric learning. Also, there are commonalities with some methods for video-based representation learning, in general, and  self-supervised learning in particular. The relation to existing methods, gaps and proposed solutions to them are discussed in this section.

\subsection{UDA with Deep Metric Learning}
The distance metric learning approach has been extensively applied to the computer vision  area, so that examples belonging to the same label (within-class samples) are close as possible in some embedding space, and samples from different labels (between-class samples) are as far from one another as possible. Recently, triplet  and Siamese networks were employed for metric learning which have been successfully applied for few-shot learning and learning with few data \cite{Hoffer15}\cite{Laradji2018}. Designing a discriminant and robust distance metric requires abundant of labeled data, and accordingly, UDA methods are required to adapt an existing metric to a new domain where data are unlabeled or scarce (or both cases, as for the applications discussed in this current work). 

There are a few works on UDA  for deep distance metric learning are recently proposed \cite{Laradji2018}\cite{sohn2018}. In \cite{Laradji2018}, the adversarial learning approach \cite{ganin2014} is applied to decrease the domain discrepancy between the datasets and simultaneously  a magnet loss is applied to align the class centers for the source and target embedding. This method works only    for closed-set problems (where the source and target share the label space).  In \cite{sohn2018}, on the other hand,    open-set problems can be tackled by introducing a separation loss so that different source and target sets are separated in the embedding space. 
  
We argue that, for such specific case of UDA (i.e., where the adapted model is a distance metric rather than feature-based model),  the   embedding  should be optimized in the distance space rather than in the feature space. In other words,   the straight forward objective function (when designing  an UDA algorithm for adapting a distance metric) is to distinguish between the different "distance" types (i.e., within-class and between class distances), and simultaneously it should be hard to identify the source of a distance sample (i.e., being constituted from samples coming from the source or the target domain). Also,  this objective produces metrics that  work for both close- and open-set problems, since it is concerned with the ultimate pairwise distances (rather than with  absolute feature representations  like with the existing methods \cite{Laradji2018}\cite{sohn2018}). This new concept is followed to design our proposed method.

\subsection{Video-based Face Representation Learning}
Face representations, based on still images, are usually designed by training deep CNNs, in general, and  Siamese CNNs, in particular. Such techniques, however, can provide unreliable performance when applied to design video-based face representations. Accordingly,  more complex models were proposed to provide improved performance, with the expense of decreased efficiency and that can hinder the real-time applicability of the designed representations. More importantly, state-of-the-art methods  require information that can be unavailable during operation (or they can be expensive to obtain) and that can make these methods impractical, especially to efficiently calibrate existing models for new video sources (i.e., cameras) using a few unlabeled data from new target subjects. 

For instance, In \cite{Ding18}, reliable detection of facial landmarks is required and that may fail due to occlusion. Also, besides the complex ensemble structure that can hinder the real-time processing, the method involves a fine tuning step that requires large amount of data from the operational target domain. In \cite{Parchami17b}, Haar-like features are extracted so that facial landmark extraction is no longer required, and in \cite{Parchami17} a lighter network structure  is proposed for improved efficiency. These methods, however, still require synthetic generation of video-like face images and  fine tuning using considerable amount of data from the target domain.  Some methods like in \cite{Wen2018}, require labeled data from the target domain for tuning.

A more recent trend is to employ self-Supervised Learning to automatically label the target data by levering  temporal and contextual information exist in videos, e.g., tracklets \cite{Sharma19}\cite{Wu13}\cite{Cinbis15}.  Although provide reasonable performance, these methods mostly require large unlabeled data to train representations from scratch. Importantly, these methods require availability of co-occurring tracklets (tracklets from different subjects in the same scene) so that negative samples can be obtained to constitute triplets. The proposed method works even with singleton tracklets  (only tracklets from a single subject appear in a scene), or where the face detector and tracker are not reliable to produce labeled tracklets.

\section{Dual-Triplet Metric Learning}
\begin{figure*}[t!]
    \centering
    \begin{subfigure}[b]{0.5\textwidth}
        \centering
        \includegraphics[width=1.0\textwidth]{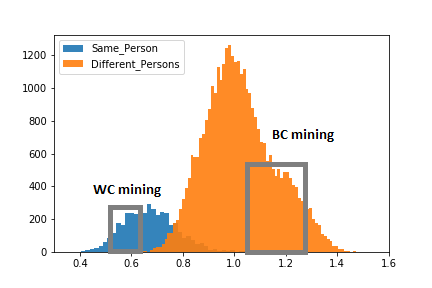}
        \caption{Source: represented by the initial source embedding}
        
    \end{subfigure}%
    ~ 
    \begin{subfigure}[b]{0.5\textwidth}
        \centering
        \includegraphics[width=1.0\textwidth]{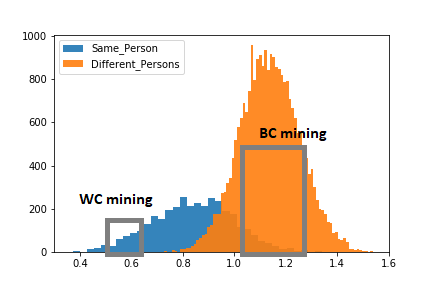}
        \caption{Target: represented by the initial source embedding}
    \end{subfigure}

    \begin{subfigure}[b]{0.5\textwidth}
        \centering
        \includegraphics[width=1.0\textwidth]{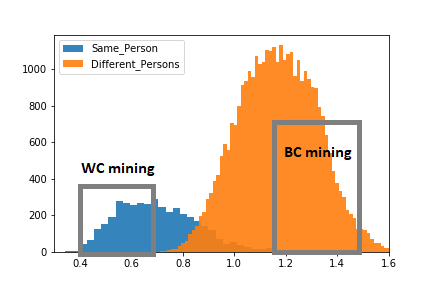}
        \caption{Source: represented by a shared embedding through DTML}
    \end{subfigure}%
    ~ 
    \begin{subfigure}[b]{0.5\textwidth}
        \centering
        \includegraphics[width=1.0\textwidth]{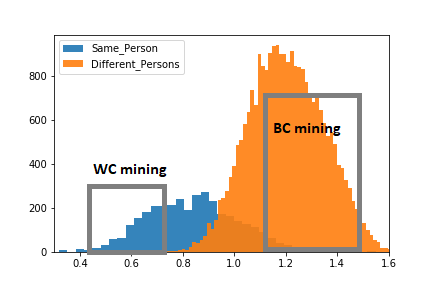}
        \caption{Target: represented by a shared embedding through DTML}
    \end{subfigure}
    \caption{Illustration of the Mutual Supervision learning using the distributions of distances between samples from  same-person  (within-class  (WC) samples) and  distances between samples from  different-persons (between-class (BC) samples). Left column (Fig.a,c) show the distance distributions for the source data (teacher) and right column (Fig.b,d) show the distributions for the target data (student). Upper row (Fig.a,b) show the distance distributions for the initial source representation, while the bottom row (Fig.c,d) show the distributions where a shared target representation is  learned using the proposed dual-triplet and mutual-supervision learning are employed.}
    \label{fig:MS}
\end{figure*}

Figure \ref{fig:DTML} illustrates the proposed Dual-Triplet Metric Learning (DTML) framework for Domain Adaptation.  This framework facilitates the calibration of a new  video source (camera) when added to an existing operational video-based face system (e.g., a video surveillance Network). Moreover, the proposed framework can be employed to adapt an existing model that works for a specific surveillance network to be operational in a different environment or even within a completely new or a different network. 

The calibration data consist of some Regions of Interest (ROIs) with faces of unknown people extracted from videos captured by the new camera. The capturing conditions of the environment and the capturing device are represented in the extracted ROIs and used to calibrate a video-based face representation of an existing video source.  The Source can be considered as a teacher as it provides the target (student) with the initial knowledge (embedding) it acquires through supervised learning, and also they (the teacher and the student) continue to learn a shared knowledge (embedding) using their different data (labeled data of the source and unlabeled calibration data of the target).

To this end, the source labeled data are used to learn an initial representation with employing the ordinary triplet-loss approach \cite{Schroff15}.  For each new (target) camera or environment, the initial source embedding is loaded  to the target and gets improved with minimizing a dual-triplet loss.     In order to constitute the target part of the dual-triplet loss,  a "mutual-supervised" process is employed, where  pairwise distances between target calibration samples are computed and  statistics of the pairwise source distances are used to label the target distances (as being within-class or between-class distances). During training, the  source and target pairwise distance distributions become more similar over time and that implies the following: 1) a distance metric that works for the source  also works for the target, and 2) the resulting metric can label pairwise distances from the target as accurate as for the source distances.

\subsection{Dual-Triplet Loss}

The dual-triplet loss $L$ consists of two terms: a source term $L_s$ and a target term $L_t$:

\begin{equation}
L=L_{s}+\lambda . L_{t}.
\end{equation} 
where  $\lambda$ is the  parameter that balancing the two objectives. 

\vskip 0.3 cm

A source triplet $L_{s}$ is constituted from labeled source data, using an anchor $a$, a positive sample $p$, a negative sample $n$, and a margin $\alpha$: 
\begin{equation}
L_{s}=max(||f(a)-f(p)||-||f(a)-f(n)||+\alpha,0)
\end{equation}
To constitute a target triplet, it can be impossible to use the absolute representation of  anchor, positive and negative samples, since the target data can be unlabeled. To resolve this limitation, we are only interested in labeling the pairwise distances as being either within-class $(wc)$ or between-class $(bc)$. Once the distances are labeled, the target triplet is constituted as follows:

\setlength{\abovedisplayskip}{1pt}
\begin{equation}
L_{t}=max(||wc||-||bc||+\alpha,0)
\end{equation}
\setlength{\belowdisplayskip}{0pt} 

The source triplet aims at designing a discriminant distance metric since it is based on labeled data from the source domain. The target triplet aims at separating the within-class and between-class target distances with the same margin as that for the source distances, so that the pairwise distance distributions of both domains are similar (and hence, the resulting distance metric  is valid for both domains). 

It is important to mention that the proposed DTML method can be applied when a few labeled samples are available from the target domain (e.g., where co-occurring tracklets are available or when some calibration data are manually annotated). In that case, the target triplets can be  constituted directly from the labeled  samples as that with the source triplets.  In case no labeled data are available from the target domain or  when only singleton tarcklets data are available (i.e., frames belong to a single person appear in the scene, so no dissimilar pairwise labels are provided by tracklets), the following mutual-supervised method is required.

\subsection{Mutual Supervision}

The objective of the mutual-supervised learning method is leverage the  labeled samples of the source to extract positive (within-class) and negative (between-class)  pairwise distance samples from the unlabeled target samples. 

Figure \ref{fig:MS} illustrates the proposes mutual-supervised learning method. Firstly, both source and target training samples are represented using the initial embedding trained using the labeled source data in supervised mode. Then, pairwise distances from both datasets are generated by computing the Euclidean distance between the feature vectors of each pair. Since the source dataset is labeled, it is straightforward to label the source pairwise distances as within-class (WC) or between-class (BC)  if they belong to same-person or different persons, respectively. Distributions of the WC and BC pairwise distance samples are generated (see  Figure \ref{fig:MS}.a). Statistics of these distributions are used to identify two mining windows: 1) within-class mining window ($WC_{mw}$) and 2) between-class mining window ($BC_{mw}$):

\begin{equation}
WC_{mw} =[\mu_{wc}-\sigma_{wc},\mu_{wc}].
\end{equation} 
\begin{equation}
BC_{mw} =[\mu_{bc}, \mu_{bc}+\sigma_{bc}].
\end{equation} 

where  $\mu_{wc}, \sigma_{wc}$ and  $\mu_{bc},\sigma_{bc}$ are the mean and standard deviation of the WC and BC pairwise distances, respectively.

\hspace*{1.0cm}

These mining windows are computed to achieve a trade-off between confidence of labeling (picking distance samples that are close enough to the center of the distributions and far from the confusion areas where WC and BC distances can overlap) and also to avoid picking too much easy  samples (samples that exist  towards the tail of the distributions as these samples  most likely lie beyond the margin so they do not contribute to the loss function).

Once the mining windows are computed based on the source pairwise distance distributions, they are used to locate (label) the target distance samples (see  Figure \ref{fig:MS}.b). Initially, the source and target distributions are not aligned (as a result of the domain shift), so using the source mining windows maybe not accurate enough and also may locate a small number of samples. These samples are used to constitute the target triplet loss term, then dual triplet is used to train a new target representation.

The above process is repeated for each training batch and eventually the source and target pairwise distance distributions become more aligned as WC and BC samples from both domains are enforced to be separated by the same margin, and a shared representation is used to represent samples from both domains.  

Figure \ref{fig:MS} (c and d) illustrate how the source and target distributions are getting aligned through the DTML with mutual-supervised training. When better aligned, the source mining Windows produce more accurate pseudo-labels of the target pairwise distance and also locate larger number of samples from each bucket (the WC and BC buckets).  Also, it is important to note that the inaccuracies of the target loss term (as a result of the imperfect pseudo-labeling) can be compensated by the existence of the perfect source loss term. 

\section{Experimental Methodology}
Although the proposed DTML framework and mutual-supervised learning approach may be applied for different modalities, we assess the methods here for the specific  still-to-video (S2V) face recognition (FR) application. To this end,  face ROIs are captured  by video cameras and matched against high quality  frontal still face images of some users enrolled to the system.

Two video-based face recognition datasets are used for the experimentation: 1) the public COX face dataset and 2) a private video-based dataset that we created internally for performance assessment. 

The COX dataset is used to assess the proposed approach ability to adapt a model that works for an existing camera in a surveillance network to be operational for a new camera added to the network. On the other hand, the private dataset  is utilized to simulate the case where a model designed for am exiting surveillance network is leveraged and adapted for a different network or operational environment. 

For the COX dataset where the standard experimental protocol described in \cite{Huang15} is followed in this experimental study, so results can be compared to the state-of-the-art methods. The dataset consists of 1000 subjects with still images are captured for each subject and then each person is captured by 3 video cameras with different views. As described in \cite{Huang15}, samples from 300 subjects are considered a training set, and the remaining 700 subjects are used for the testing. To simulate the camera calibration scenario, we split the training set to 200 subjects for training the initial source models, and 100 subjects for calibrate the new camera. This split is important to simulate the case where subjects used during developing the initial solution (i.e., in the lab) are different than the subjects who appear in the operational field during a calibration session. This setup also simulate the "open-set" scenario, where subjects appear during operation are not seen during the design and calibration phases.  

The private video-based dataset consists of 100 subjects where only video-based face images are captured by a commercial IP video surveillance camera during real operational setup. Since there is no still templates are collected during operation, we manually selected best face image from each subject (i.e., nearly frontal, best size and quality, etc) and used them as a still images gallery. 
The training set consists of 30 subjects and the testing set consists of the remaining 70 subjects. 
Since we simulate the case where a model from an existing network is leveraged and adapted for a new network or environment, so here we use a model tuned for camera 1 from the COX dataset as a source, and employ the proposed approach to adapt this work for the camera that we used to create our private dataset. Accordingly, the whole training set (30 subjects) are used for calibration.

For all experiments,  samples are firstly represented using  the VGG Face representation \cite{Cao18}. Then, a source embedding is trained using triplet loss. This embedding is used to test the case where only source models are leveraged without any domain adaptation step. To this end, the S2V performance for the  subjects of the testing set is computed and considered as a lower-bound performance.

To simulate the camera calibration process using our proposed UDA approach, the calibration and training data (from source and target cameras, respectively) are represented by the learned source representation and used to constitute the dual-triplet terms. Then, the DTML algorithm runs to learn an embedding for the calibrated camera.

To simulate the case where labeled data are available from the calibrated camera (e.g., through expensive manual annotation), the DTML is employed but the target labels are taken directly from the dataset instead of the labels produced by the mutual-supervised method.

To test the impact of the dual-triplet loss terms, three scenarios are implemented:

\begin{enumerate}
 \item $L_s$: where only data from the source camera are used and only the the source triplet is used to tune the network. This scenario simulates the case where we don't use calibration data and only keep improving the source representation.
 
 \item $L_t$: where only data from the target camera are used and only the the target triplet is used to tune the network. This scenario simulates the case where we only rely on calibration data to adapt the source model for the new camera. 
 
 \item $L_s +L_t$: where data from both  the source and calibrated cameras are used for adaptation, which is the exact DTML proposed method.
\end{enumerate}

For all experiments, the batch size is set to 100 (5 persons per batch, 20 images per person). The source labels are used to load training images from five different persons per batch, while the proposed mutual-supervision method is employed to pseudo-label pair-wise image distances so that equivalent number of similar and dissimilar images are loaded from the target dataset.  Then, DTML runs for 40 epochs, and performance is tested using rank 1 accuracy and the Area Under ROC Curves (AUC). Parameter $\lambda$ (see Eq. 1) is used to balance the contributions of the source and target triplet terms to the overall loss function. Extensive experimentation show that equal contributions lead to best results, so we set $\lambda=1$ in all reported results.

\section{Experimental Results and Discussion}

\begin{table*}
\centering
\begin{tabular}{l|l|l|l|l|l|l|}
\cline{2-7}

 & \multicolumn{2}{l|}{C3  $\rightarrow$   C1}  & \multicolumn{2}{l|}{C1 $\rightarrow$ C3}   & \multicolumn{2}{l|}{C3 $\rightarrow$ C2}\\ \hline

\multicolumn{1}{|l|}{\textbf{Methods}} & AUC & Acc   & AUC & Acc   & AUC & Acc  \\ \hline

\multicolumn{1}{|l|}{VGG-Face \cite{Cao18}}
& - & 69.6
& - & 68.1
& - & 76.0

\\ \hline\hline

\multicolumn{1}{|l|}{TBE-CNN \cite{Ding18}}
& - & 87.8
& - & 88.2
& - & 95.7
\\ \hline 
\multicolumn{1}{|l|}{CCM-CNN \cite{Parchami17}}
& - & 87.8
& - & 88.6
& - & 92.1
\\ \hline 
\multicolumn{1}{|l|}{HaarNet \cite{Parchami17b}}
& - & 87.9
& - & 89.3
& - & 97.0

\\ \hline\hline
\multicolumn{1}{|l|}{\begin{tabular}[c]{@{}l@{}}Source model: without DA\end{tabular}} 
& 0.98 & 95.0
& 0.90 & 81.7
& 0.94 & 87.2
\\ \hline  
\multicolumn{1}{|l|}{\begin{tabular}[c]{@{}l@{}}Proposed UDA: DTML-A\end{tabular}} 
& 0.99 & 97.7
& 0.95 & 88.7
& 0.97 & 91.0

\\ 
\hline

\multicolumn{1}{|l|}{\begin{tabular}[c]{@{}l@{}}Upper-bound: supervised DA\end{tabular}} 
& 0.99 & 98.0
& 0.98 & 93.3
& 0.99 & 95.3

\\ \hline 

\end{tabular}
\caption{Performance of the proposed DTML method with mutual supervision on the COX dataset.}
\label{table:results}
\end{table*}

\begin{figure}[thpb]
      \centering
      \includegraphics[scale=0.21]{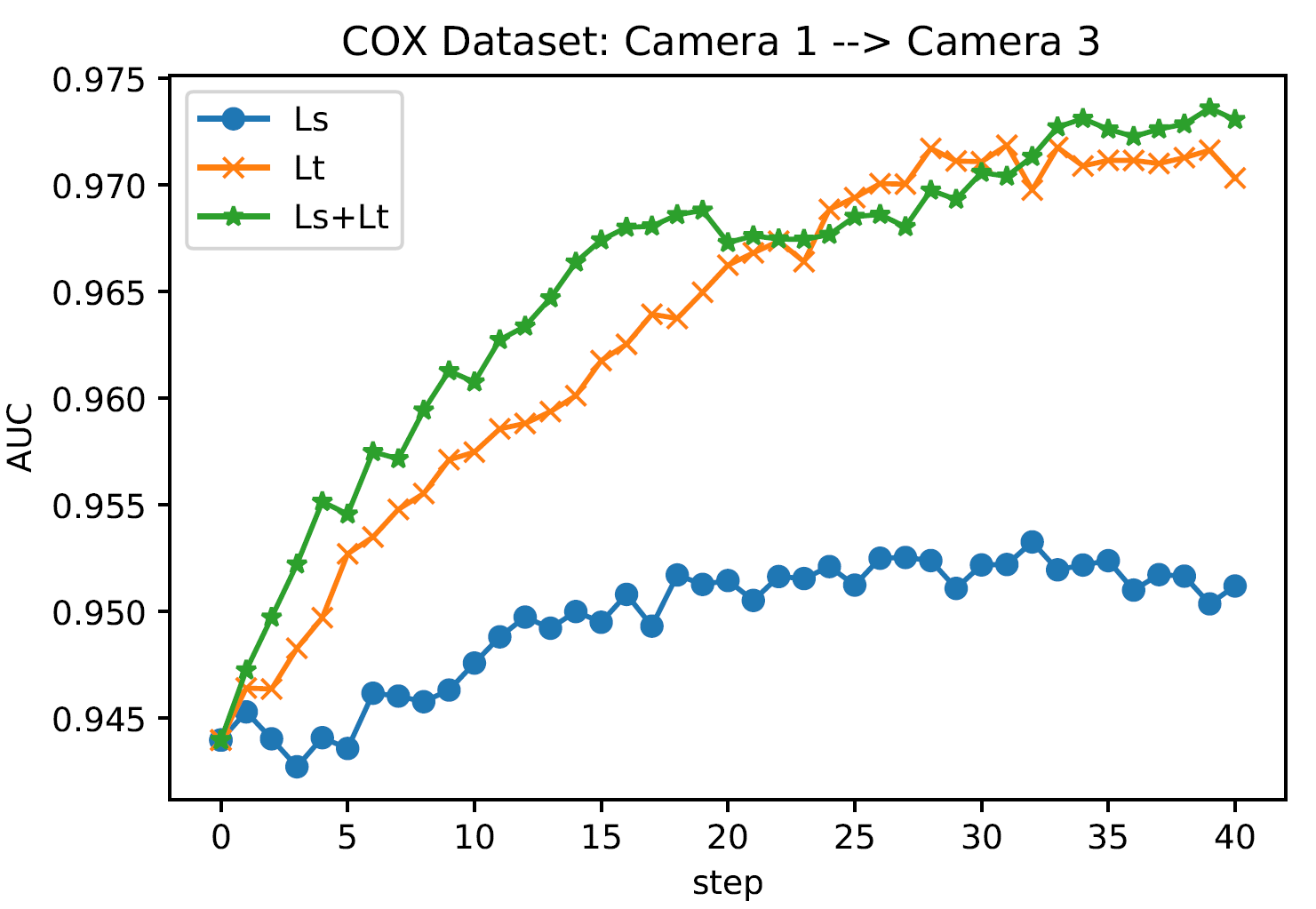}
      \caption{COX dataset: Camera 1 to Camera 3}
      \label{fig:cam1_cam3}
   \end{figure}

\begin{figure}[thpb]
      \centering
      \includegraphics[scale=0.21]{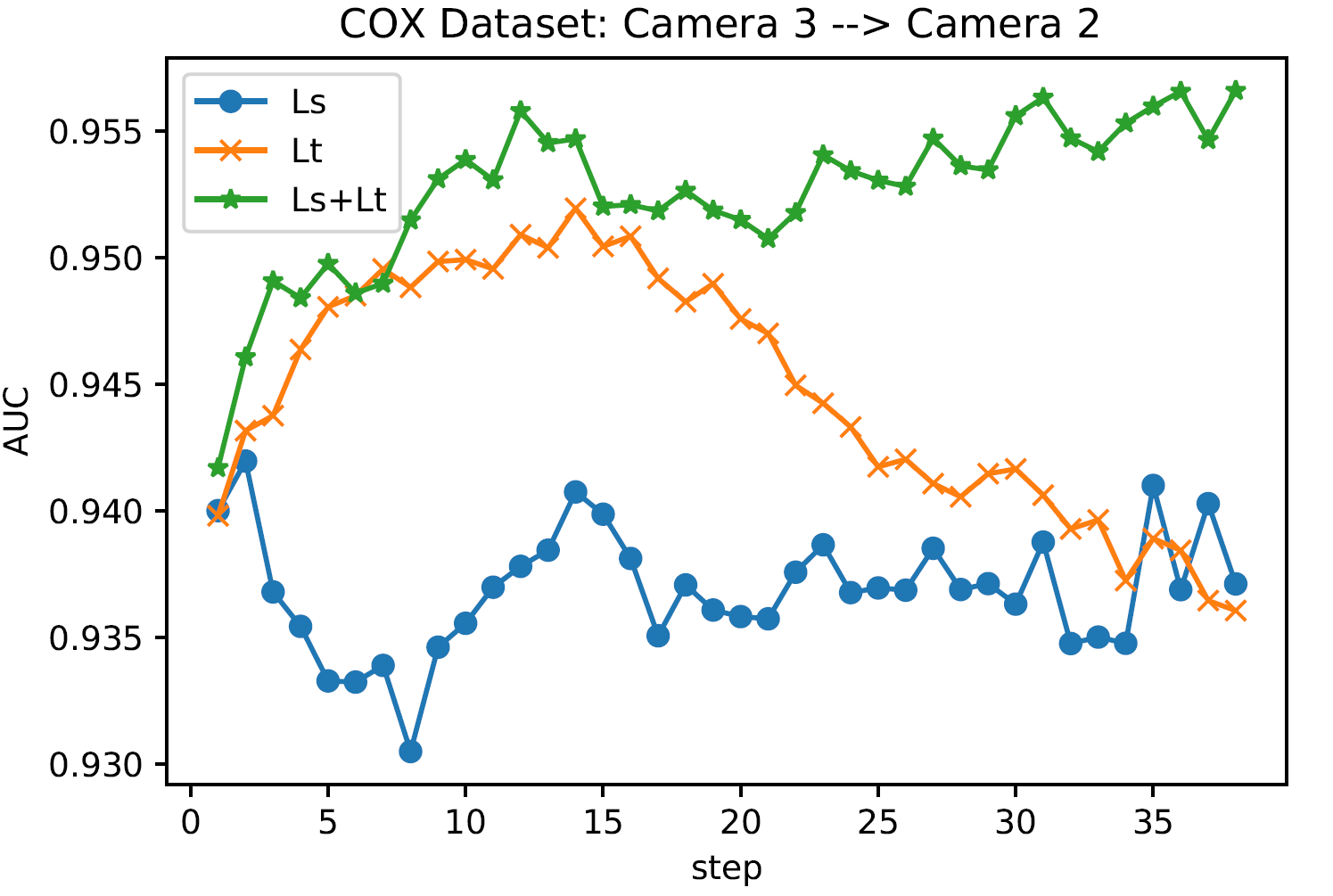}
      \caption{COX dataset: Camera 3 to Camera 2}
      \label{fig:cam3_cam2}
   \end{figure}
\begin{figure*}[t!]
    \centering
    \begin{subfigure}[b]{0.5\textwidth}
        \centering
        \includegraphics[width=0.9\textwidth]{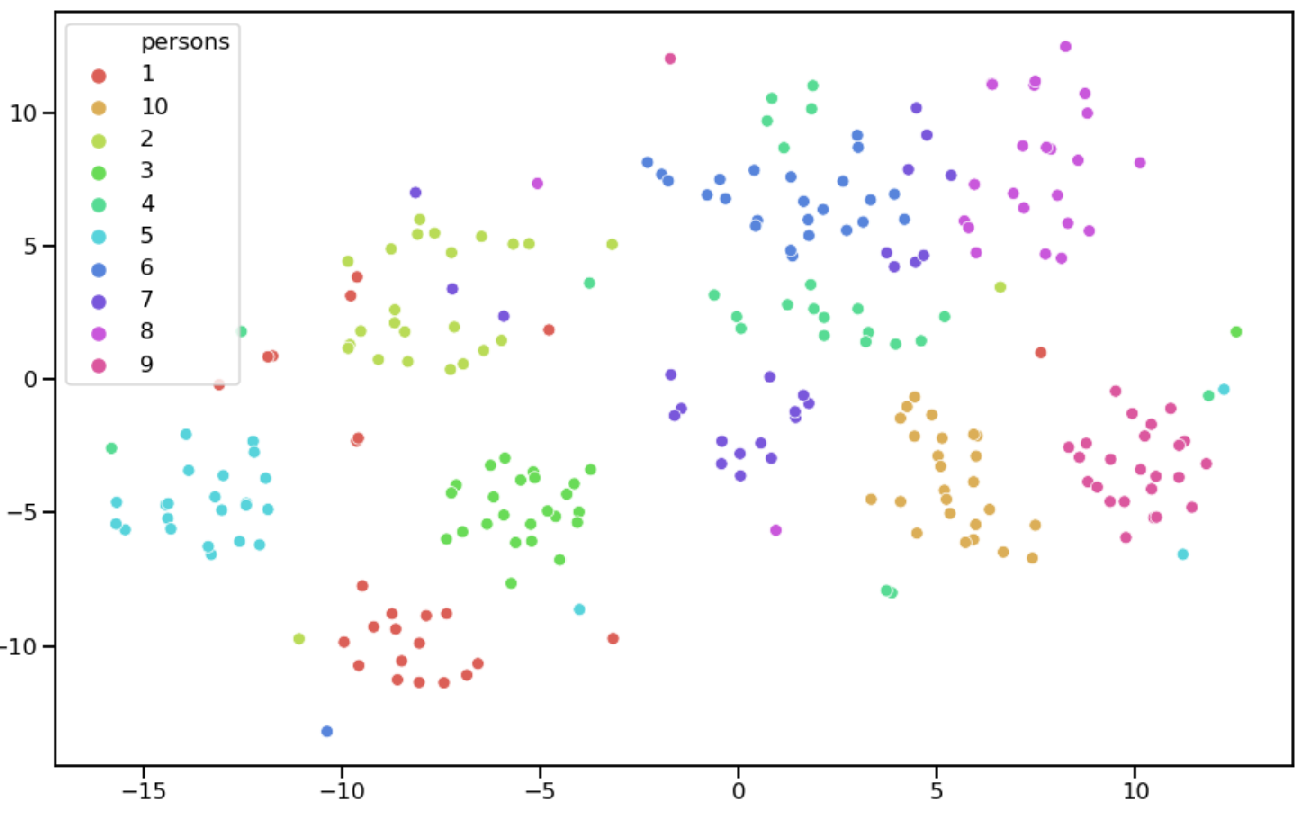}
        \caption{Before domain adaptation}
    \end{subfigure}%
    ~ 
    \begin{subfigure}[b]{0.5\textwidth}
        \centering
        \includegraphics[width=0.9\textwidth]{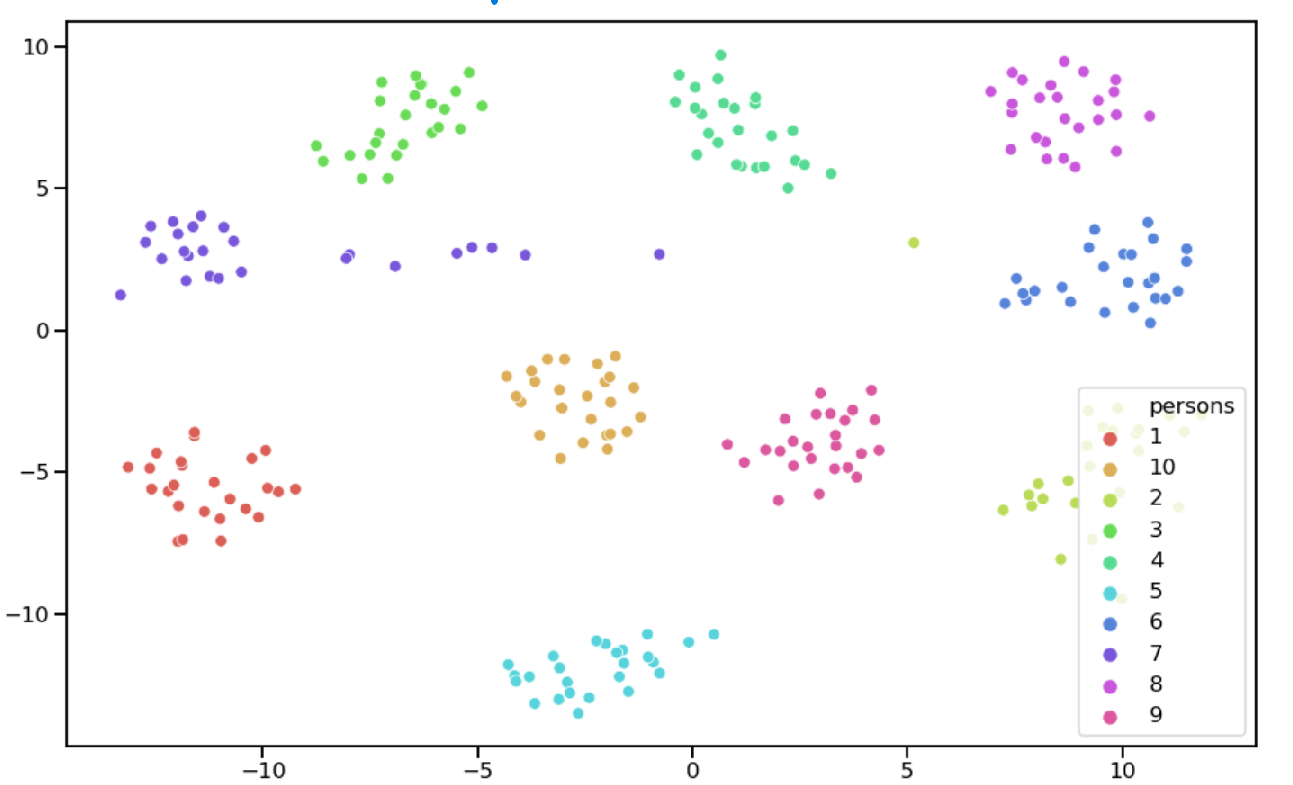}
        \caption{After domain adaptation.}
    \end{subfigure}
    
    \caption{COX dataset: TSNE representation for first ten persons in the testing set}
       \label{fig:TSNECOX}

\end{figure*}

\begin{figure*}[t!]
    \centering
    \begin{subfigure}[b]{0.5\textwidth}
        \centering
        \includegraphics[width=0.9\textwidth]{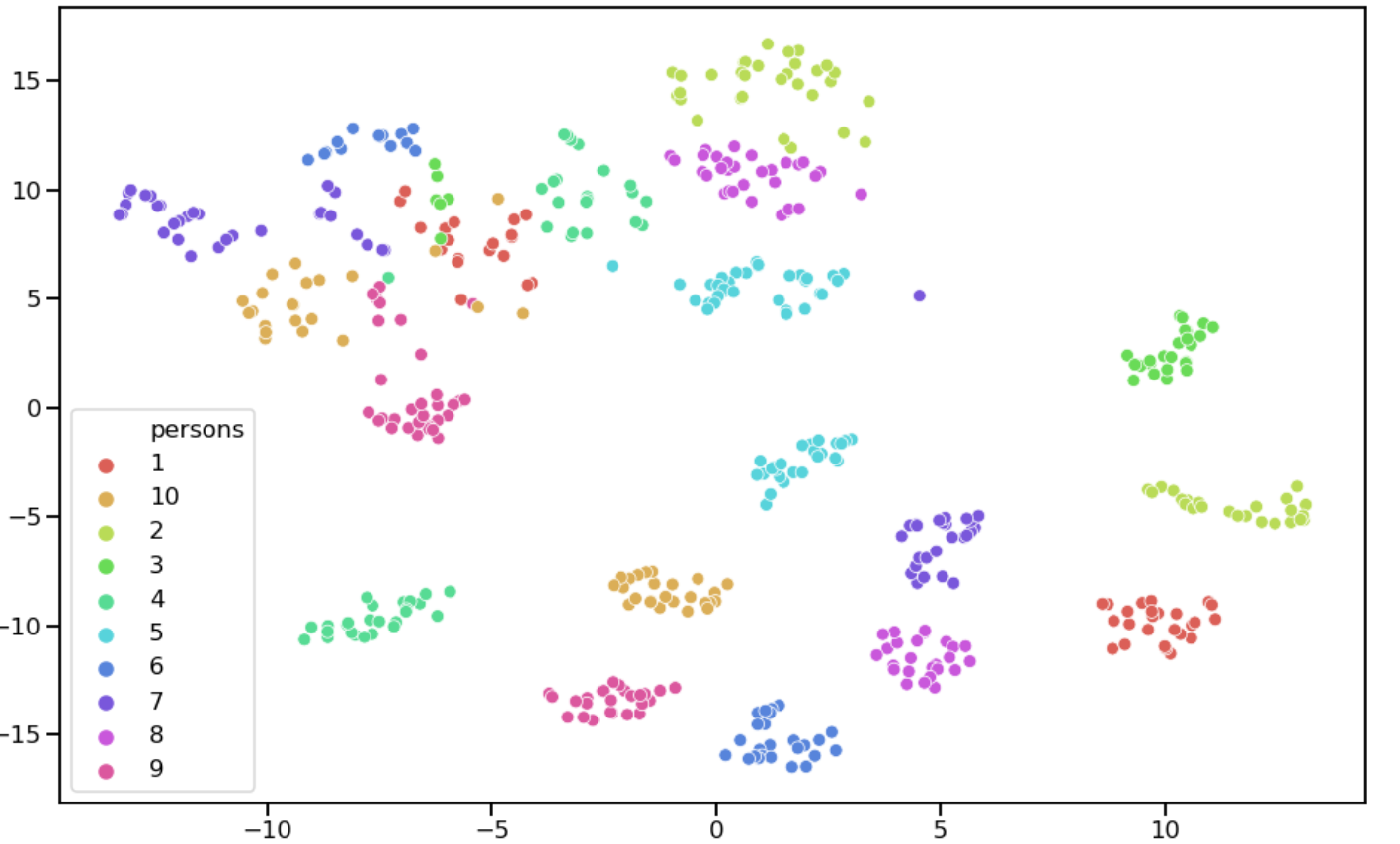}
        \caption{Before domain adaptation}
    \end{subfigure}%
    ~ 
    \begin{subfigure}[b]{0.5\textwidth}
        \centering
        \includegraphics[width=0.9\textwidth]{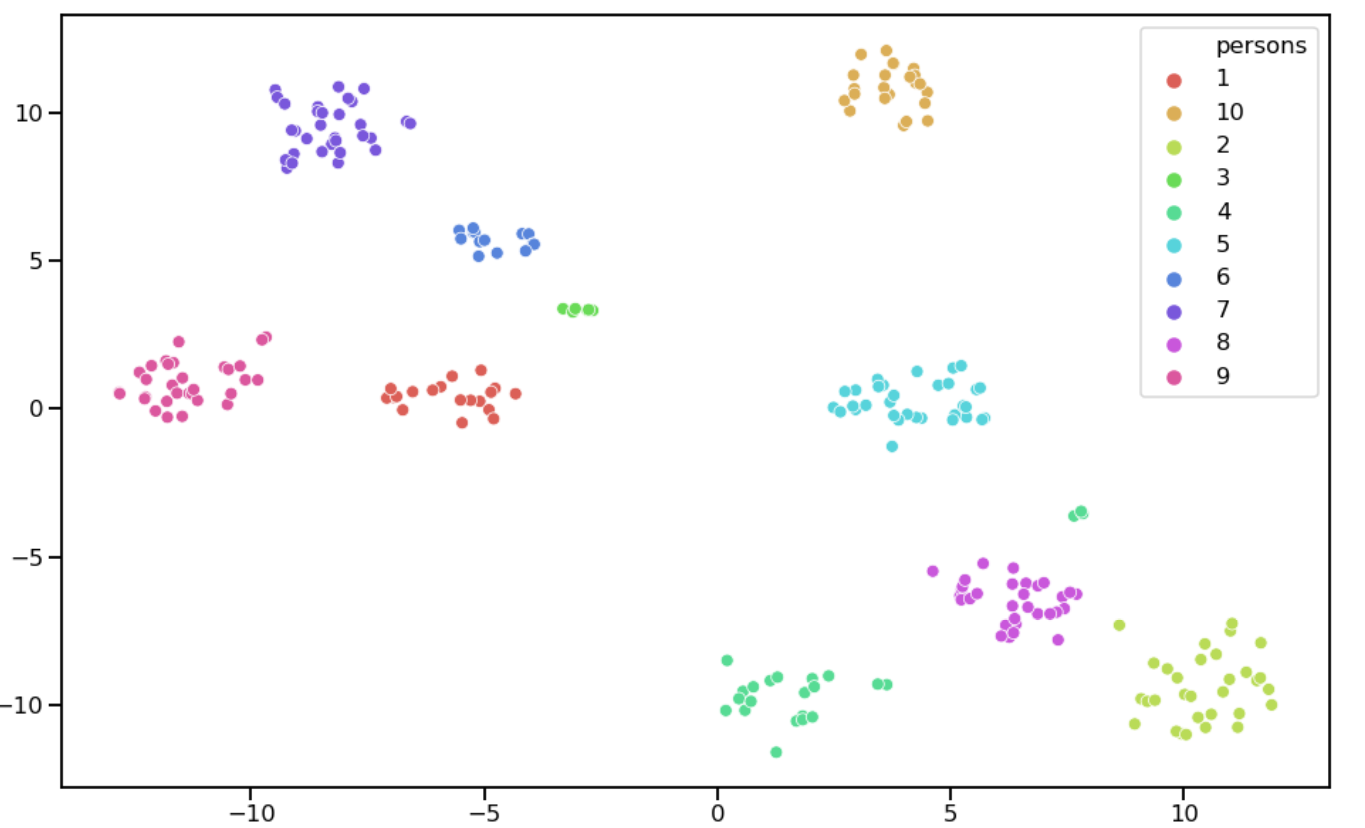}
        \caption{After domain adaptation}
    \end{subfigure}

    \caption{Private Video Face dataset: TSNE representation for first ten persons in the testing set}
    \label{fig:TSNEPRIVATE}
\end{figure*}

Table \ref{table:results} shows the S2V Face Recognition results using the COX dataset.  It is obvious that tuning the basic feature representation (VGG-Face) using the source data only provides improved accuracy, even without applying any domain adaptation step. This observation is expected, as cameras within same network have some similarities, and accordingly training for one camera can help for the other cameras.

Applying the proposed DTML algorithm with mutual-supervision achieved a significant improvement for both domain shift problems (around 5\% increase in AUC for both cases). While the source model performance is very low compared to the upper-bound performance (in case that labels of the  calibration data are available, e.g., through expensive manually annotation, perfect co-occurring tracklets, etc.), the proposed method could perform a reliable domain adaptation in an unsupervised fashion (i.e., UDA) and achieved a performance close to the upper-bound.

Comparing with the performance of state-of-the-art methods, the proposed method achieves comparable results (for the Cam3 $\rightarrow$ Cam2 case), and somewhat less performing (for the Cam1 $\rightarrow$ Cam3 case). Noting that these methods are either less efficient (e.g., because  they involve complex ensembles of CNNs, expensive generation of synthetic video-like ROIs, etc.), or they rely on information that can be unavailable  during camera calibration (e.g., facial marks that can be unavailable due to occlusion, abundant of calibration data, etc.). On the other hand, the proposed  DTML  with mutual-supervision approach employs simple CNN structures, and only uses a small amount of unlabeled calibration data without the need to generate synthetic captures or extract facial landmarks.

Figures \ref{fig:cam3_cam2} and \ref{fig:cam1_cam3} illustrates the impact of the dual triplet terms. In Figure \ref{fig:cam3_cam2}, it is clear that only using the source data (i.e., using loss triplet term $L_s$ only) is not helpful to classify samples from the target camera. Also, using the target loss ($L_t$) alone, although gained performance increase during the initial training iteration, the performance is dramatically decreased  afterwards due to model overfitting. Having both  the source and target triplets ($L_s + L_t$) guarantee continuous performance increase over the training period.

In Fig \ref{fig:cam1_cam3}, it is clear that source data alone provides a limited performance gain. Similar to the above observation,  the best performance is achieved when triplets from both domains are optimized. Although using triplets from only the target domain provided adequate performance for this domain shift problem, this result wouldn't be achieved without having the help of the source as the target data are initially unlabeled and they are labeled using the source embedding (in order to constitute the target triplets).

Fig \ref{fig:TSNECOX} illustrates the impact of applying the proposed UDA method to adapting a  model designed for an existing camera to a new camera so that surveillance networks are extensible. Fig \ref{fig:TSNECOX}.a shows the TSNE representation of the testing samples from the first ten subjects captured by camera 2 (target camera, newly added to the network) of the COX dataset, where the embedding is generated using a model tuned for camera 3 (source camera, already exist in the network). It is clear that the source embedding is not suitable enough for the target camera. Fig \ref{fig:TSNECOX}.b shows the the TSNE representation of the target camera after adapting the model using the proposed UDA approach. It is clear that the adapted model provides discriminant representation.

Fig \ref{fig:TSNEPRIVATE} illustrates the impact of applying the proposed method to leveraging a model of an existing network to be function for a new network or environment where collecting and labeling enough  data from the new network to train a model from scratch  is infeasible. Fig \ref{fig:TSNEPRIVATE}.a shows the TSNE representation of the testing samples from the first ten subjects captured by the target camera (a commercial IP camera used to capture video-based face image in a realistic operation conditions in an uncontrolled environment). The embedding is generated using a model that is fine tuned for camera 1 from the COX dataset. Although we have chosen the source camera from the COX dataset (that is closest to the target camera as it has the least domain shift between source and target), it is obvious most clusters (subjects representation) are split into two separate sets and there is a significant overlap between subjects(see the top part of the plot). When we applied the proposed UDA approach, using unlabeled samples from the target camera, the clusters are well separated (see \ref{fig:TSNEPRIVATE}.b) and accordingly the FR accuracy has significantly improved (True positive rate at False Alarm rate (FAR=1\%) increased from 52\% from 73\%).

Figure \ref{fig:MS} (that is used to illustrate the proposed mutual-supervised learning method in in Section III.B; these distance distribution plots are generated using camera 1 from the COX dataset as a source and our private video-face dataset as a target)  shows the separability of WC and BC distance distributions  of the target data where the source initial  embedding is adapted using the proposed UDA approach. With UDA,  the source and target distributions are well aligned. More specifically, after adaptation,  a simple threshold, e.g., 0.9 provides a good trade-off between false positives and false negatives for both of the source and target domains, as apposed to to initial state (Figures \ref{fig:MS} (a and b) where the source and target distance distributions are not aligned and a threshold that works for the source (e.g., 0.8) results in high FAR rates when used by the target. 

It is important to mention that, although the adapted embedding provides more aligned and separable pairwise distance distributions (i.e.,  target within-class (WC) and between-class (BC) distributions are better separated and aligned with the source distributions), it is not expected to rely only on the quality of resulting embedding  to provide accurate classification results when simple Euclidean distance is used as a distance metric, and also when a simple threshold is used for classification in the embedding space.

Accordingly, we further feed the embedding of the still (template) and video
(query) samples to a two-layer fully connected network and we trained this outer network separately using the pseudo-labeling approach in Section III.B. This step has improved the recognition accuracy from $73\%$ to  $84\%$ with $FAR=1\%$. More importantly,  instead of feeding the two streams of embedding (from both the still and video samples), we further generate the following dissimilarity feature representation:

\begin{equation}
\delta(X) =|X^{Q} - X^{T}|.
\end{equation} 

where $X^{Q}$ and $X^{T}$ are the feature representation generated by embedding adapted with the proposed DTML approach for the query (video) and template (still) samples, respectively. The resulting dissimilarity  representation is accordingly of the same dimensionality  as that for the original representation. We noticed  a significant improvement in accuracy due this transformation step (recognition accuracy has increased from  $84\%$ to  $90\%$ with $FAR=1\%$). Future work will explore  employing the DTML approach with having the dissimalrity representation and the outer layers trained in an end-to-end fashion.

Also, it is important to note that the proposed method is only applicable to transfer problems where the domain shift between the source and target domain is small enough so that mutual-supervised training is possible. For instance, if  the initial  source and target distance distribution are not aligned enough so that mining approach locate completely wrong samples, or even does not any sample from either the WC or the BC buckets, so in this case the mutual-supervised learning mechanism will not work correctly (see Figure \ref{fig:MS}).  Future work will explore different methods to enforce the distance distributions of the different domains to be aligned without the need for the pseudo-labeling step, by employing the adversarial learning concept.

\section{Conclusion}
A general method for domain adaptation of deep distance metrics is proposed in this paper.  The proposed method is applicable to domain shift problems where the target domain can provide a small amount of data, and also the cases where  only unlabeled  target data are available.  Two main aspects of the proposed method are discussed: the dual-triplet optimization and the mutual-supervision process. Having  a dual-triplet form the source and target domains helps in avoiding model corruption because if the lack of target data, and also makes the pairwise distance distributions of both domains more similar. The mutual-supervision feature provides a tool to constitute triplets from the unlabelled target samples. The method is applied  to provide  unsupervised domain adaptation  for the still-to-video face recognition systems and achieved a level of accuracy comparable to the state-of-the-art more complex methods, that also can be impractical given the limitations of  the camera calibration use-case.

{\small
\bibliographystyle{ieee}

\begin{thebibliography}{21}
\providecommand{\natexlab}[1]{#1}
\providecommand{\url}[1]{\texttt{#1}}
\expandafter\ifx\csname urlstyle\endcsname\relax
  \providecommand{\doi}[1]{doi: #1}\else
  \providecommand{\doi}{doi: \begingroup \urlstyle{rm}\Url}\fi

\bibitem[{Cinbis} et~al.(2011){Cinbis}, {Verbeek}, and {Schmid}]{Cinbis15}
R.~G. {Cinbis}, J.~{Verbeek}, and C.~{Schmid}.
\newblock Unsupervised metric learning for face identification in tv video.
\newblock In \emph{2011 International Conference on Computer Vision}, pp.\
  1559--1566, Nov 2011.
\newblock \doi{10.1109/ICCV.2011.6126415}.

\bibitem[de~Amorim \& Hennig(2015)de~Amorim and Hennig]{Renato15}
Renato~Cordeiro de~Amorim and Christian Hennig.
\newblock Recovering the number of clusters in data sets with noise features
  using feature rescaling factors.
\newblock \emph{Inf. Sci.}, 324:\penalty0 126--145, 2015.

\bibitem[Ding \& Tao(2018)Ding and Tao]{Ding18}
C.~Ding and D.~Tao.
\newblock Trunk-branch ensemble convolutional neural networks for video-based
  face recognition.
\newblock \emph{IEEE transactions on pattern analysis and machine
  intelligence}, 40\penalty0 (4):\penalty0 1002--1014, 2018.

\bibitem[Ganin \& Lempitsky(2014)Ganin and Lempitsky]{ganin2014}
Yaroslav Ganin and Victor Lempitsky.
\newblock Unsupervised domain adaptation by backpropagation.
\newblock \emph{arXiv}, stat.ML\penalty0 (1409.7495v2), September 2014.

\bibitem[Hoffer \& Ailon(2015)Hoffer and Ailon]{Hoffer15}
Elad Hoffer and Nir Ailon.
\newblock Deep metric learning using triplet network.
\newblock In Aasa Feragen, Marcello Pelillo, and Marco Loog (eds.),
  \emph{Similarity-Based Pattern Recognition}, pp.\  84--92, Cham, 2015.
  Springer International Publishing.
\newblock ISBN 978-3-319-24261-3.

\bibitem[Hong et~al.(2017)Hong, Im, Ryu, and Yang]{hong2017}
Sungeun Hong, Woobin Im, Jongbin Ryu, and Hyun~S. Yang.
\newblock {SSPP-DAN}- deep domain adaptation network for face recognition with
  single sample per person.
\newblock \emph{arXiv}, cs.CV\penalty0 (1702.04069v4), February 2017.

\bibitem[Huang et~al.(2015)Huang, Shan, Wang, Lao, Kuerban, and Chen]{Huang15}
Z.~Huang, S.~Shan, R.~Wang, S.~Lao, A.~Kuerban, and X.~Chen.
\newblock A benchmark and comparative study of video-based face recognition on
  cox face database.
\newblock \emph{IEEE Transactions on Image Processing}, 14\penalty0
  (12):\penalty0 5967--5981, 2015.

\bibitem[Laradji \& Babanezhad(2018)Laradji and Babanezhad]{Laradji2018}
Issam~H. Laradji and Reza Babanezhad.
\newblock M-adda: Unsupervised domain adaptation with deep metric learning.
\newblock \emph{CoRR}, abs/1807.02552, 2018.

\bibitem[{Luo} et~al.(2018){Luo}, {Hu}, {Deng}, and {Shen}]{Luo2018}
Z.~{Luo}, J.~{Hu}, W.~{Deng}, and H.~{Shen}.
\newblock Deep unsupervised domain adaptation for face recognition.
\newblock In \emph{2018 13th IEEE International Conference on Automatic Face
  Gesture Recognition (FG 2018)}, pp.\  453--457, May 2018.
\newblock \doi{10.1109/FG.2018.00073}.

\bibitem[Mokhayeri et~al.(2019)Mokhayeri, Granger, and
  Bilodeaun]{Mokhayeri2019}
F~Mokhayeri, E~Granger, and GA~Bilodeaun.
\newblock Domain-specific face synthesis for video face recognition from a
  single sample per person.
\newblock \emph{IEEE Transactions on Information Forensics and Security},
  14\penalty0 (3), 2019.

\bibitem[Parchami et~al.(2017{\natexlab{a}})Parchami, Bashbaghi, and
  Granger]{Parchami17}
M.~Parchami, S.~Bashbaghi, and E.~Granger.
\newblock Cnns with cross-correlation matching for face recognition in video
  surveillance using a single training sample per person.
\newblock In \emph{Proceedings of the 14th IEEE International Conference on
  Advanced Video and Signal Based Surveillance (AVSS)}, pp.\  1--6,
  2017{\natexlab{a}}.

\bibitem[Parchami et~al.(2017{\natexlab{b}})Parchami, Bashbaghi, and
  Granger]{Parchami17b}
M.~Parchami, S.~Bashbaghi, and E.~Granger.
\newblock Video-based face recognition using ensemble of haar-like deep
  convolutional neural networks.
\newblock In \emph{Proceedings of the International Joint Conference on Neural
  Networks (IJCNN)}, pp.\  4625--4632, 2017{\natexlab{b}}.

\bibitem[Parchami et~al.(2017{\natexlab{c}})Parchami, Bashbaghi, Granger, and
  Sayed]{Parchami17c}
M.~Parchami, S.~Bashbaghi, E.~Granger, and S.~Sayed.
\newblock Using deep autoencoders to learn robust domain-invariant
  representations for still-to-video face recognition.
\newblock In \emph{Proceedings of the 14th IEEE International Conference on
  dvanced Video and Signal Based Surveillance (AVSS)}, 2017{\natexlab{c}}.

\bibitem[{Q. Cao} et~al.(2018){Q. Cao}, {Shen}, {Xie}, {Parkhi}, and
  {Zisserman}]{Cao18}
Q.~{Q. Cao}, L.~{Shen}, W.~{Xie}, O.~M. {Parkhi}, and A.~{Zisserman}.
\newblock Vggface2: A dataset for recognising face across pose and age.
\newblock In \emph{2018 International Conference on Automatic Face and Gesture
  Recognition}, 2018.

\bibitem[Schroff et~al.(2015)Schroff, Kalenichenko, and Philbin]{Schroff15}
F.~Schroff, D.~Kalenichenko, and J.~Philbin.
\newblock Facenet: A unified embedding for face recognition and clustering.
\newblock In \emph{Proceedings of the Proceedings of the IEEE conference on
  computer vision and pattern recognition (CVPR)}, pp.\  815--823, 2015.

\bibitem[Sharma et~al.(2019)Sharma, Tapaswi, Sarfraz, and
  Stiefelhagen]{Sharma19}
Vivek Sharma, Makarand Tapaswi, M.~Saquib Sarfraz, and Rainer Stiefelhagen.
\newblock Self-supervised learning of face representations for video face
  clustering.
\newblock \emph{2019 14th IEEE International Conference on Automatic Face and
  Gesture Recognition (FG 2019)}, pp.\  1--8, 2019.

\bibitem[Sohn et~al.(2017)Sohn, Liu, Zhong, Yu, Yang, and Chandraker]{Sohn2017}
Kihyuk Sohn, Sifei Liu, Guangyu Zhong, Xiang Yu, Ming-Hsuan Yang, and Manmohan
  Chandraker.
\newblock Unsupervised domain adaptation for face recognition in unlabeled
  videos.
\newblock \emph{arXiv}, cs.CV, cs.AI\penalty0 (1708.02191v1), August 2017.

\bibitem[Sohn et~al.(2019)Sohn, Shang, Yu, and Chandraker]{sohn2018}
Kihyuk Sohn, Wenling Shang, Xiang Yu, and Manmohan Chandraker.
\newblock Unsupervised domain adaptation for distance metric learning.
\newblock In \emph{International Conference on Learning Representations}, 2019.

\bibitem[Wen et~al.(2018)Wen, Chen, Cai, and He]{Wen2018}
Ge~Wen, Huaguan Chen, Deng Cai, and Xiaofei He.
\newblock Improving face recognition with domain adaptation.
\newblock \emph{Neurocomputing}, 287:\penalty0 45 -- 51, 2018.
\newblock ISSN 0925-2312.
\newblock \doi{https://doi.org/10.1016/j.neucom.2018.01.079}.
\newblock URL
  \url{http://www.sciencedirect.com/science/article/pii/S0925231218301127}.

\bibitem[{Wolf} et~al.(2011){Wolf}, {Hassner}, and {Maoz}]{Wolf11}
L.~{Wolf}, T.~{Hassner}, and I.~{Maoz}.
\newblock Face recognition in unconstrained videos with matched background
  similarity.
\newblock In \emph{CVPR 2011}, pp.\  529--534, June 2011.
\newblock \doi{10.1109/CVPR.2011.5995566}.

\bibitem[{Wu} et~al.(2013){Wu}, {Lyu}, {Hu}, and {Ji}]{Wu13}
B.~{Wu}, S.~{Lyu}, B.~{Hu}, and Q.~{Ji}.
\newblock Simultaneous clustering and tracklet linking for multi-face tracking
  in videos.
\newblock In \emph{2013 IEEE International Conference on Computer Vision}, pp.\
   2856--2863, Dec 2013.
\newblock \doi{10.1109/ICCV.2013.355}.

\end{thebibliography}

}

\end{document}